\documentclass[letterpaper, 10 pt, conference]{ieeeconf}  

\IEEEoverridecommandlockouts                              
\let\labelindent\relax

\usepackage{graphics} 
\usepackage{graphicx}
\usepackage{epsfig} 
\usepackage{mathptmx} 
\usepackage{times} 
\usepackage{amsmath} 
\usepackage{amssymb}  
\usepackage{hyperref}

\usepackage{amsthm}
\usepackage{dsfont}
\usepackage{xcolor}
\usepackage[shortlabels]{enumitem}

\usepackage{mathtools}
\theoremstyle{definition}
\newtheorem{definition}{Definition}[section]

\DeclarePairedDelimiter\floor{\lfloor}{\rfloor}
\newcommand\scaleddot{\scalebox{.89}{.}}
\makeatletter
\renewcommand{\dddot}[1]{%
  {\mathop{\kern\z@#1}\limits^{\makebox[0pt][c]{\vbox to-2.2\ex@{\kern-\tw@\ex@
  \hbox{\normalfont\scaleddot\kern-0.5pt\scaleddot\kern-0.5pt\scaleddot}\vss}}}}}
\renewcommand{\ddddot}[1]{%
  {\mathop{\kern\z@#1}\limits^{\makebox[0pt][c]{\vbox to-2.2\ex@{\kern-\tw@\ex@
  \hbox{\normalfont\scaleddot\kern-0.5pt\scaleddot\kern-0.5pt\scaleddot\kern-0.5pt\scaleddot}\vss}}}}}
\makeatother

\title{\LARGE \bf Controlling the Cascade: Kinematic Planning for $N$-ball Toss Juggling}
\author{Kai Ploeger and Jan Peters
\thanks{K. Ploeger and J. Peters are with the Technical University of Darmstadt.}}

\begin{document}

\maketitle
\thispagestyle{empty}
\pagestyle{empty}



\begin{abstract}
Dynamic movements are ubiquitous in human motor behavior as they tend to be more efficient and can solve a broader range of skill domains than their quasi-static counterparts.
For decades, robotic juggling tasks have been among the most frequently studied dynamic manipulation problems since the required dynamic dexterity can be scaled to arbitrarily high difficulty.
However, successful approaches have been limited to basic juggling skills, indicating a lack of understanding of the required constraints for dexterous toss juggling.
We present a detailed analysis of the toss juggling task, identifying the key challenges of the switching contacts task and formalizing it as a trajectory optimization problem.
Building on our state-of-the-art, real-world toss juggling platform, we reach the theoretical limits of toss juggling in simulation, evaluate a resulting real-time controller in environments of varying difficulty and achieve robust toss juggling of up to 17 balls on two anthropomorphic manipulators. \href{https://sites.google.com/view/controlling-the-cascade}{https://sites.google.com/view/controlling-the-cascade}
\end{abstract}

\section{Introduction}
The task of juggling pushes the boundary of control performance of robotic manipulators, testing both their speed and dexterity.
In this paper, we consider the task of toss juggling to investigate the feasibility of achieving $N$-ball juggling in the intricate cascade patterns achieved by skilled humans.
By pushing the boundary of dynamic robotic manipulation with juggling, we aim to stimulate research for faster and more dexterous robot control, as juggling requires consideration of both mechanical design and control strategy.
By opening up the solution space of possible control strategies away from popular static approaches to more dynamic alternatives~\cite{mason1993dynamic, furukawa2006dynamic, ha2022flingbot}, we gain potential benefits in overcoming torque constraints and under-actuation, energy efficiency, speed, simpler movements, and simpler hardware requirements.

\subsection{Problem Statement}
The definition of juggling is controversial both in the juggling and in the scientific community.
We restrict our study to single person toss juggling of balls by four assumptions.

\begin{definition}[Simple Toss Juggling]
~
\begin{enumerate}[a)]
    \item Balls are held in hand for a non-zero duration;
    \item Balls are thrown targeting the same height;
    \item Balls are thrown from and aimed at defined locations;
    \item Balls are thrown to a constant rhythm.
\end{enumerate}
\label{def:toss_juggling}
\end{definition}

\noindent While Definition~\ref{def:toss_juggling} is a simplification compared to most definitions of toss juggling, it does not make the task easy. On the contrary, instead of just requiring no ball to drop to the ground, it is our goal to closely track ball trajectories that are well defined through take-off and touch-down location as well as flight time or throw height.

\newpage

\begin{figure}
    \centering
    \vspace{5pt} 
    \includegraphics[width=1\columnwidth, trim={0cm, 0cm , 0cm, 0cm}, clip]{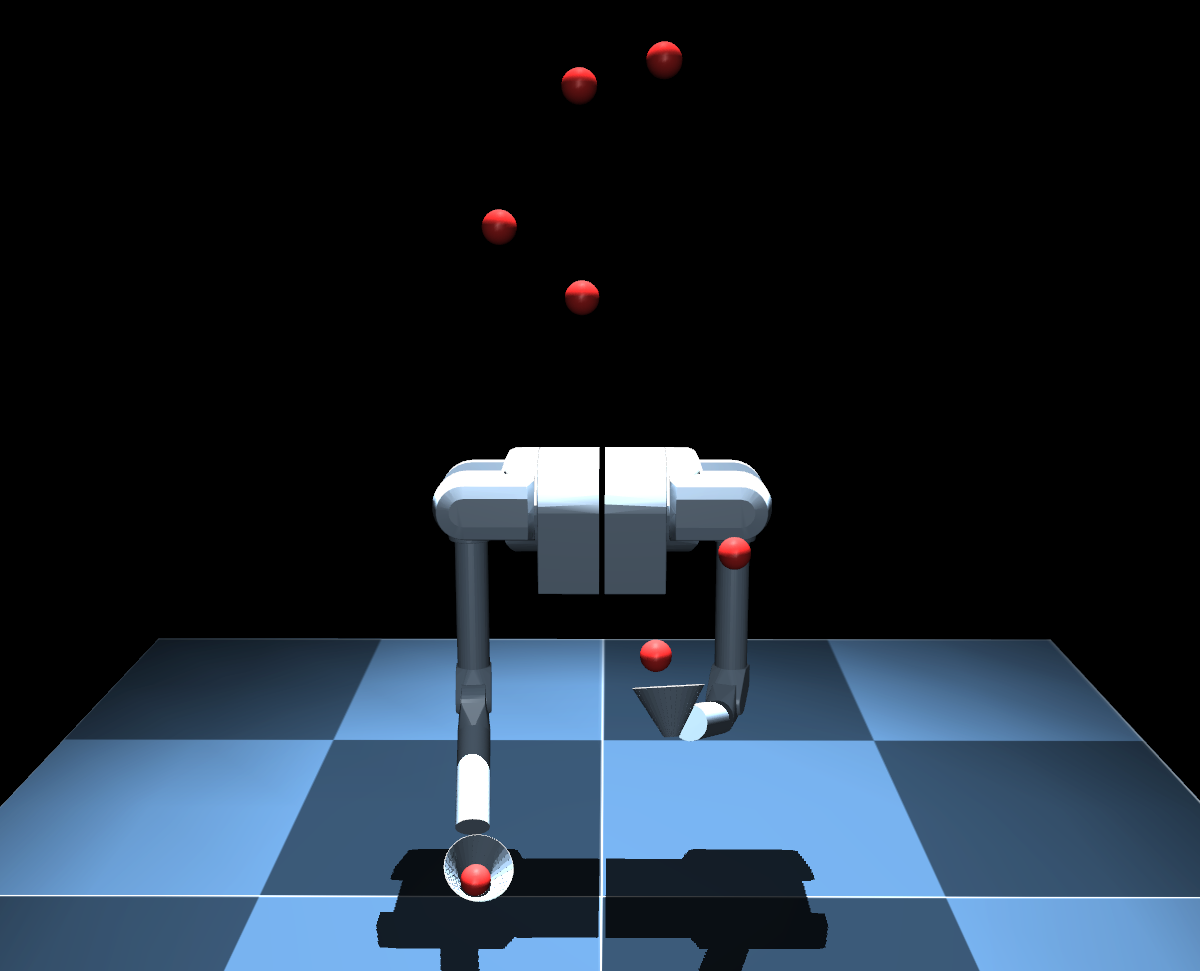}
    \caption{Stable juggling of seven and more balls in a cascade pattern on an anthropomorphic two-arm robotic setup. Trajectories are planned in real-time and adapted to disturbances.}
    \vspace{-10pt}
    \label{fig:title_pic}
\end{figure}

Given Definition~\ref{def:toss_juggling}, two distinct juggling patterns arise: For odd numbers of balls, they always cross from one hand to the other following the cascade pattern, and for even numbers of balls, they always return to the same hand following the fountains pattern illustrated in figure~\ref{fig:patterns}.
Juggling a cascade pattern with an even number of balls or a fountain pattern with an odd number of balls is not possible without breaking the symmetry of the movement or ball collisions.
In both kinds of patterns, balls have to be caught at and thrown from sufficiently distant locations to avoid ball collisions.

Juggling is an infinite horizon problem that can be divided into a sequence of finite-horizon problems alternating between catching and throwing.
The main problem in catching is to decide where to catch the object. In juggling, the search space is reduced by aiming balls to intercept a desired touch-down location.
Also, the difficulty of the throwing task is reduced by aiming each throw at the same target.
Why is juggling more challenging than either of its parts? Because each sub-task in the sequence effects the next task, leading to quickly compounding errors.
The catching phase can only start after clearing the previous ball, leaving just a fraction of the flight time of each ball to execute the catch.
The limited time a ball spends in-hand leaves little time to recover from the distribution of catching positions and requires a matching distribution of throwing movements starting at each catching position.
As these time constraints can be changed arbitrarily by adjusting the targeted pattern, toss juggling is an excellent subject to explore dynamic manipulation.

\newpage
\subsection{Toss Juggling Theory}
\begin{figure}[t]
    \centering
    \vspace{5pt}
    \includegraphics[width=1\columnwidth, trim={0cm, 0cm , 0cm, 0cm}, clip]{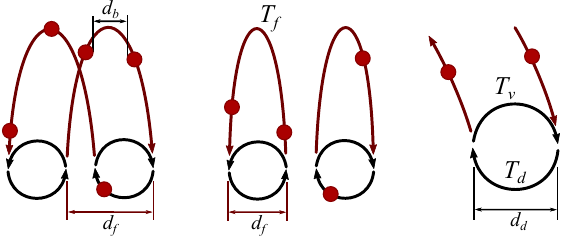}
    \vspace{-10pt}
    \caption{(Left) Crossing throws in a five ball \emph{cascade} (Middle) Non-crossing throws in a four-ball \emph{fountain} pattern (Right) The handcycle with dwell time $T_d$ and vacant time $T_v$ and carry distance $d_d$.}
    \vspace{-10pt}
    \label{fig:patterns}
\end{figure}
As shown in figure~\ref{fig:patterns}, each hands cycle time $T_c=T_v+T_d$ is split into a vacant time $T_v$ that it spends empty and a dwell time $T_d$ holding a ball. The catching movement is executed during $T_v$ and the throwing movement during $T_d$. Both can be related to the flight time $T_f$ that each ball spends in the air by Claude Shannon's juggling theorem~\cite{shannon1993scientific, yam1998extending}
\begin{equation*}
   \frac{T_f + T_d}{N_b} = \frac{T_v+T_d}{N_h},
   \label{eq:shannon}
\end{equation*}
where $N_b$ is the number of juggled balls, and $N_h$ is the number of hands. The dwell ratio $R={T_d}/{T_c}$, $R \in (0,1)$ describes how long balls stay in a hand, with typical values for humans around $R\approx 2/3$.
Increasing the dwell ratio reduces the average amount of balls in the air per hand
\begin{equation*}
    W=\frac{T_f}{T_c}=\frac{N_b}{N_h}-R.
\end{equation*}
Assuming that the horizontal velocities of all balls on the same parabola are constant and identical, we can relate the horizontal distance between two subsequently thrown balls 
\begin{equation*}
    d_b = \frac{d_f}{W} - 2r_b
\end{equation*}
to the distance $d_f$ between take-off and touch-down locations and the ball radii $r_b$.
The horizontal distance between balls is also their minimal distance that is reached near the top of the pattern at equal height. Requiring the minimal distance $d_b$ to be positive results in a kinematic upper bound on $N_b$ that is independent of the throw height.
\begin{equation}
    N_b < \floor*{\frac{d_f}{r_b} + 2R} \label{eq:limit}
\end{equation}
In fountain patterns $d_f$ only depends on the carry distance $d_d$ and can not be increased by moving the hands further apart. Therefore, they are generally considered more difficult than cascade patterns by requiring larger hand movements.
\vspace{10pt}

\noindent
\textbf{Contributions}:
We propose a novel decomposition of the infinite-horizon toss juggling movement into a sequential short-horizon trajectory optimization problem, identifying the critical constraints necessary for this dexterous dynamic manipulation task with switching contacts.
To the best of our knowledge, this formulation is the first to demonstrate stable juggling of five and more balls with anthropomorphic manipulators
in a physics based environment.

\newpage
\section{Related Work}

%

From a mathematic perspective, toss juggling has been studied exhaustively~\cite{shannon1993scientific, buhler1994juggling, beek1995science, yam1998extending, polster2003mathematics, mays2006combinatorial}, while roboticists focused mostly on the related task of paddle juggling that has no dwell time.
The robot-ball interactions can then be modeled as impulse exchanges \cite{buehler1987robotics, aboaf1987task} leading to a class of mirror algorithms~\cite{buhler1990family, buehler1994planning} that were extended to two balls on a spatial robot~\cite{rizzi1992progress, rizzi1993further}.
Open-loop stability of paddle juggling has been achieved by decelerating during the upward stroke~\cite{schaal1993open, schaal1996one} and a parabolic paddle~\cite{reist2012design}.


The two components of toss juggling, throwing and catching, have been studied individually.
%
%
%
While throwing, dynamic effects such as dynamic closure~\cite{mason1993dynamic} and whipping in the kinematic chain~\cite{senoo2008high} are at play.
Most approaches capable of reliably hitting targets have been data-driven, by iterative adaptation of the aiming~\cite{aboaf1988task} or by optimizing a movement primitive through trial and error~\cite{kober2012playing}.
%

The necessary complexity of approaches for the ball-catching task depends on the mechanical setup. In case of contacts with low restitution---balls not bouncing off the hand---it is sufficient to intercept the ball trajectory.
Online trajectory optimization was successfully applied minimizing the mechanical energy~\cite{lampariello2011trajectory} or various kinematic cost functions~\cite{bauml2010kinematically}.
A common simplification to the catching problem is the introduction of a catching plane~\cite{kober2012playing}.
In case of higher restitution, it is necessary to reduce impact velocities to prevent the ball from bouncing off.
Additionally, matching the accelerations~\cite{hove1991experiments, namiki2014ball} widens the time interval of low relative velocity, increasing robustness to disturbances.
To solve the more complex task of catching non-spherical objects, distributions over grasp positions and catching behaviors were learned from demonstrations~\cite{kim2014catching}.

%

Claude Shannon did the first investigations in robotic toss juggling in the form of juggling mechanisms solving a particular case of toss juggling, where balls are bounced off a drum on the floor.
While mechanisms of this kind~\cite{schaal1993open, atkeson2017shannon} implemented the carry distance $d_d$ necessary to avoid collisions by rolling the balls in the hand, later juggling mechanisms~\cite{machines1002007juggling} also implemented an active carry phase.
All juggling mechanisms relied on funnel-shaped hands dissipating excess energy for passive stability.

Most throwing strategies in robotic juggling are open-loop as well.
An elliptic trajectory~\cite{sakaguchi1991study} was executed on a robot with two degrees of freedom (DoF). Initializing and improving the trajectory parameterization through a ballistic model led to stable one-handed two-ball juggling.
The addition of a hand-tuned throwing motion to a previous catching approach~\cite{kober2012playing} resulted in a shared three ball cascade with a human partner.
Another approach also used a hand-tuned throwing motion~\cite{kizaki2012two}, but in this case, the ball was caught by the same articulated hand-arm system that threw the ball to juggle two balls in one hand.
In our previous work~\cite{ploeger2021high}, we used model-free episodic reinforcement learning on the physical system to optimize a movement primitive and achieved stable one-handed two-ball toss juggling for up to 33 minutes on an anthropomorphic manipulator.

\newpage
\section{Juggling in Task Space}
Dexterous toss juggling can be formulated as a trajectory optimization problem.
Since hands do not interact directly, the movements of each hand can be planned individually but need to be conditioned on the states of the balls. 

\subsection{Throwing}

For precise throwing, a set of take-off constraints has to hold. The take-off hand position $\mathbf{x}(t_{\mathrm{TO}})$ and hand velocity $\dot{\mathbf{x}}(t_{\mathrm{TO}})$ at $t_{\mathrm{TO}}$ have to result in a target touch-down ball location $\mathbf{b}(t_{\mathrm{TD}+1})=\mathbf{b}_{\mathrm{TD,des}}$ after the flight time ${T_f=t_{\mathrm{TD}+1}-t_{\mathrm{TO}}}$. According to the parabolic ballistic model ${\mathbf{b}(t)= \mathbf{b}_{\mathrm{TO}}(t) + \dot{\mathbf{b}}_{\mathrm{TO}}(t)t + \frac{1}{2}}\mathbf{g}t^2$ we define our take-off constraints similar to~\cite{mason1993dynamic} as
\begin{align}
    \mathbf{0} &= \mathbf{x}(t_{\mathrm{TO}}) - \mathbf{b}_{\mathrm{TO,des}}, \label{eq:TO_pos}\\
    \mathbf{0} &= \mathbf{x}(t_{\mathrm{TO}}) + \alpha\dot{\mathbf{x}}(t_{\mathrm{TO}})T_f + \frac{1}{2}\mathbf{g}T_f^2 - \mathbf{b}_{\mathrm{TD,des}}, \label{eq:TO_vel}\\
    \mathbf{0} &= \ddot{\mathbf{x}}(t_{\mathrm{TO}}) - \mathbf{g}, \label{eq:TO_acc}
\end{align}
assuming ${\mathbf{b}_{\mathrm{TO}}=\mathbf{x}_{\mathrm{TO}}}$ and $\dot{\mathbf{b}}_{\mathrm{TO}}=\alpha\dot{\mathbf{x}}_{\mathrm{TO}}$. 
The take-off is defined as that moment in time at which the contact forces between ball and hand vanish, which is ensured by constraint~(\ref{eq:TO_acc}).
The scalar ${\alpha \approx 1}$ compensates for the small amount of potential energy that is stored in the ball-hand contact and would lead to slight excess ball take-off velocity ${\lVert \dot{\mathbf{b}}_{\mathrm{TO}} \rVert > \lVert \dot{\mathbf{x}}_{\mathrm{TO}} \rVert}$.
To prevent post-take-off ball contacts with parts of the hand by premature lateral movement, we require the relative acceleration between ball and hand to be collinear with the hand normal vector $\mathbf{e}_h$ on a short interval ${\tau \in (0, T_{\mathrm{PTO}}]}$ after take-off, as shown in figure~\ref{fig:constraints}
\begin{equation}
    \mathbf{0} = (\ddot{\mathbf{x}}(\tau) - \mathbf{g}) \times \mathbf{e}_{h}.
\end{equation}
Since planned cycles end at $t_{\mathrm{TO}}$, this constraint acts on the initial interval of the next planned movement.

\subsection{Catching}

To catch a ball, we predict the time ${\tilde{t}_{\mathrm{TD}}}$ and location ${\tilde{\mathbf{b}}_{\mathrm{TD}}}$ of touch-down on a fixed horizontal plane and match the hand position ${\mathbf{x}(\tilde{t}_{\mathrm{TD}})}$ accordingly
\begin{equation}
    \mathbf{0} = \mathbf{x}(\tilde{t}_{\mathrm{TD}}) - \tilde{\mathbf{b}}_{\mathrm{TD}}.
    \label{eq:TD_pos}
\end{equation}
To prevent pre-touch-down contact between the ball and the outer side of the hand, the relative velocity between ball and hand should be collinear with the hand normal vector on a short interval ${\tau \in [0, T_{\mathrm{PTO}}]}$ before touch-down by
\begin{equation*}
    \mathbf{0} = (\dot{\mathbf{x}}(\tilde{t}_{\mathrm{TD}}-\tau) - \tilde{\dot{\mathbf{b}}}(\tilde{t}_{\mathrm{TD}}-\tau)) \times \mathbf{e}_{h},
\end{equation*}
analogous to constraint~(\ref{eq:TO_acc}) but in velocities, not in accelerations. Since we found this constraint challenging to optimize, we approximate it efficiently through
\begin{equation}
    \mathbf{0} = \dot{\mathbf{x}}(\tilde{t}_{\mathrm{TD}}-\tau) \times \tilde{\dot{\mathbf{b}}}(\tilde{t}_{\mathrm{TD}}-\tau),
    \label{eq:TD_vel}
\end{equation}
choosing a hand normal $\mathbf{e}_h$ near collinear to the ball's predicted touch down velocity $\tilde{\dot{\mathbf{b}}}_{\mathrm{TD}}$.

\begin{figure}
    \centering
    \vspace{5pt} 
    \includegraphics[width=0.7\columnwidth, trim={0cm, 0cm , 0cm, 0}, clip]{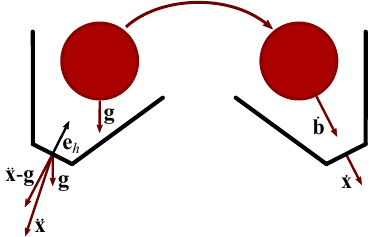}
    \vspace{-5pt}
    \caption{\textbf{Pre-touch-down and post-take-off constraints:} (Left) After take-off, the relative hand acceleration $\ddot{\mathbf{x}}-\mathbf{g}$ and the hand normal $\mathbf{e}_h$ are constrained to be collinear. (Right) Before touch-down, the hand velocity $\dot{\mathbf{x}}$ and the ball velocity $\dot{\mathbf{b}}$ are constrained to be collinear.}
    \label{fig:vis_constraints}
\end{figure}

\subsection{Trajectory Optimization}

Given a desired take-off location $\mathbf{b}_{\mathrm{TO,des}}$, position, velocity, and acceleration are fully constrained such that subsequent cycles of the juggling motion are independent of each other. Therefore, we formulate the mathematical program as a discrete-time trajectory optimization problem with piece-wise constant jerk ${\dddot{\mathbf{x}}_k \in \{\dddot{\mathbf{x}}_0, \dots, \dddot{\mathbf{x}}_{K_c-1}\}}$  over one cycle, starting at the moment of take-off $t_{\mathrm{TO}-1}$ of the previous cycle, while complying with constraints $\mathbf{h}$~(\ref{eq:TO_pos}-\ref{eq:TD_vel}).
\begin{align}
    \min_{\{\dddot{\mathbf{x}}_0, \dots, \dddot{\mathbf{x}}_{K_c-1}\}} & \sum_{k=0}^{K_c} \ddot{\mathbf{x}}_k^T \ddot{\mathbf{x}}_k \text{ s.t.} \label{eq:opt_task} \\
   (\mathbf{x}_0, \dot{\mathbf{x}}_0, \ddot{\mathbf{x}}_0) &=  (\mathbf{x}_{\mathrm{TO}-1}, \dot{\mathbf{x}}_{\mathrm{TO}-1}, \ddot{\mathbf{x}}_{\mathrm{TO}-1})\nonumber \\
   \mathbf{h}(\mathbf{x}_{k}, \dot{\mathbf{x}}_{k}, \ddot{\mathbf{x}}_{k}, \mathbf{x}_{k}) &= \mathbf{0} \nonumber\\
   -\dddot{x}_{i,\mathrm{max}} &\leq \dddot{x}_{k,i} \leq \dddot{x}_{i,\mathrm{max}} \nonumber
\end{align}
To avoid unattainable peak accelerations, we penalize the sum of squared accelerations and box-constrain the jerk $\dddot{\mathbf{x}}_k$ at each support point $k$ and each dimension $i$.
Planning in piece-wise constant jerk instead of piece-wise constant acceleration results in piece-wise linear acceleration.
This allows us to define a precise time for the acceleration constraint~(\ref{eq:TO_acc}) to hold despite discretizing the time.
Additionally, it allows us to increase the discretization time step $\Delta t$, reducing the number of decision variables.
We solve the problem in CasADi~\cite{andersson2019casadi} by direct shooting with interior point methods and the explicit Euler integration scheme
\begin{align}
    \mathbf{x}_{k+1} &= \mathbf{x}_{k} + \dot{\mathbf{x}}_{k}\Delta t + \frac{1}{2}\ddot{\mathbf{x}}_{k} \Delta t^2 + \frac{1}{6}\dddot{\mathbf{x}}_{k}\Delta t^3, \nonumber \\
    \dot{\mathbf{x}}_{k+1} &= \dot{\mathbf{x}}_{k} + \ddot{\mathbf{x}}_{k}\Delta t + \frac{1}{2}\dddot{\mathbf{x}}_{k} \Delta t^2, \label{eq:integration} \\
    \ddot{\mathbf{x}}_{k+1} &= \ddot{\mathbf{x}}_{k} + \dddot{\mathbf{x}}_{k}\Delta t \nonumber.
\end{align}
Since we integrate the kinematics, not the system dynamics, symplectic integration is not necessary. 
The resulting piece-wise linear acceleration allows us to plan smooth trajectories despite a low number of support points $K_c < 30$.
Additionally---in contrast to piece-wise constant accelerations---the take-off acceleration constraint~(\ref{eq:TO_acc}) is only fulfilled exactly at $t_{\mathrm{TO}}$, cleanly defining from which position $\mathbf{x}_{\mathrm{TO}}$ the ball takes off and with which velocity $\dot{\mathbf{x}}_{\mathrm{TO}}$.

\section{Juggling in Joint Space}

\subsection{Planning}

To plan juggling movements in joint space, we formulate the trajectory optimization problem similar as in task space~(\ref{eq:opt_task}) and solve for a cycle of piece-wise constant joint space jerks  ${\dddot{\mathbf{q}}_k \in \{ \dddot{\mathbf{q}}_0, \dots, \dddot{\mathbf{q}}_{K_c-1}\}}$, integrating analogous to~(\ref{eq:integration}).
\vspace{-10pt}
\begin{align}
    \min_{\{\dddot{\mathbf{q}}_0, \dots, \dddot{\mathbf{q}}_{K_c-1}\}} & \sum_{k=0}^{K_c} \ddot{\mathbf{q}}_k^T \ddot{\mathbf{q}}_k \text{ s.t.} \nonumber \label{eq:opt_joint}\\
   (\mathbf{q}_0, \dot{\mathbf{q}}_0, \ddot{\mathbf{q}}_0) &=  (\mathbf{q}_{\mathrm{TO}-1}, \dot{\mathbf{q}}_{\mathrm{TO}-1}, \ddot{\mathbf{q}}_{\mathrm{TO}-1})\nonumber \\
   \mathbf{h}(\mathbf{q}_{k}, \dot{\mathbf{q}}_{k}, \ddot{\mathbf{q}}_{k}, \mathbf{q}_{k}) &= 0 \nonumber\\
   -\dddot{q}_{i,\mathrm{max}} &\leq \dddot{q}_{k,i} \leq \dddot{q}_{i,\mathrm{max}} \nonumber
\end{align}
Before, the robot and ball states were defined in a shared euclidean space. Now the robot state is defined in the joint space. Therefore we need to reformulate constraints $\mathbf{h}$~(\ref{eq:TO_pos}-\ref{eq:TD_vel}), which condition the robot's movement on the predicted and desired ball trajectories.
First, we match the balls predicted touch-down location $\mathbf{x}_{\mathrm{TD}} = \tilde{\mathbf{b}}_{\mathrm{TD}}$ and desired take-off location $\mathbf{x}_{\mathrm{TO}} = \mathbf{b}_{\mathrm{TO}}$ as well as compute a desired take-off hand velocity by solving equation~(\ref{eq:TO_vel}) for 
\begin{equation}
\dot{\mathbf{x}}_{\mathrm{TO}} = \frac{\mathbf{b}_{\mathrm{TD+1,des}}-\mathbf{b}_{\mathrm{TO}}}{\alpha T_f} - \frac{\mathbf{g}T_f}{2\alpha}. \nonumber
\end{equation}
Since $\mathbf{x}_{\mathrm{TO}}$, $\dot{\mathbf{x}}_{\mathrm{TO}}$, $\mathbf{g}$, and $\mathbf{x}_{\mathrm{TD}}$ are independent of the decision variables, we can solve the inverse kinematics problems
\begin{align}
    (\mathbf{q}_{\mathrm{TO}}, \dot{\mathbf{q}}_{\mathrm{TO}}, \ddot{\mathbf{q}}_{\mathrm{TO}}) &= \mathbf{f}_{\mathrm{kin}}^{-1}(\mathbf{x}_{\mathrm{TO}}, \dot{\mathbf{x}}_{\mathrm{TO}}, \mathbf{g}), \nonumber \\
    \mathbf{q}_{\mathrm{TD}} &= \mathbf{f}_{\mathrm{kin}}^{-1}(\mathbf{x}_{\mathrm{TD}}), \nonumber
\end{align}
a priori. Now constraints~(\ref{eq:TO_pos}-\ref{eq:TO_acc}) and~({\ref{eq:TD_vel}}) can be replaced by
\begin{align}
    (\mathbf{q}_{k_{\mathrm{TO}}}, \dot{\mathbf{q}}_{k_{\mathrm{TO}}}, \ddot{\mathbf{q}}_{k_{\mathrm{TO}}}) &= (\mathbf{q}_{\mathrm{TO}}, \dot{\mathbf{q}}_{\mathrm{TO}}, \ddot{\mathbf{q}}_{\mathrm{TO}}), \nonumber \\
    \mathbf{q}_{k_{\mathrm{TD}}} &= \mathbf{q}_{\mathrm{TD}}. \nonumber
\end{align}
Finally, we extend the post-take-off constraints~(\ref{eq:TO_acc}) and the pre-touch-down constraints~(\ref{eq:TD_vel}) by solving the forward kinematics at optimization time, 
\vspace{-7pt}
\begin{align}
    0 &= \dot{\mathbf{x}}_{k_{\mathrm{TD}}-i} \times  \dot{\mathbf{b}}_{k_{\mathrm{TD}}-i}, \label{eq:pre_TD_vel} \\
    \dot{\mathbf{x}}_{k_{\mathrm{TD}}-i} &= \mathbf{J}(\mathbf{q}_{k_{\mathrm{TD}-i}})\dot{\mathbf{q}}_{k_{\mathrm{TD}-i}}, \label{eq:pre_TD_kin} \\
    0 &= (\ddot{\mathbf{x}}_{j} - \mathbf{g}) \times \mathbf{e}_{\mathrm{hand}}(\mathbf{q}_{j}), \label{eq:post_TO_acc} \\
    \ddot{\mathbf{x}}_{j} &= \mathbf{J}(\mathbf{q}_{j})\ddot{\mathbf{q}}_{j} +  \dot{\mathbf{J}}(\mathbf{q}_{j},\dot{\mathbf{q}}_{j})\dot{\mathbf{q}}_{j}, \label{eq:post_TO_kin}
\end{align}
with hand Jacobian $\mathbf{J}$ at support points $i\in \{1,\dots,N_{\mathrm{TD}}\}$ before touch-down and support points $j\in \{1,\dots,N_{\mathrm{TO}}\}$ after the previous take-off. The previous take-off is at $k=0$, so the post-take-off constraints apply to the start of the movement.

\subsection{Control}
To execute the planned trajectories, a tracking controller is necessary. We consider a simple proportional derivative (PD) controller~(\ref{eq:PD}) with the options to add either gravity compensation~(PD+G)~(\ref{eq:PD}) or a feed-forward torque~(PD+FF)~(\ref{eq:ID}) that is pre-computed from the planned trajectory through the inverse dynamics model $\mathbf{f}_{\mathrm{dyn}}^{-1}$.
Lastly, we test a joint space inverse dynamics controller~(ID)~(\ref{eq:ID}) that also uses the same inverse dynamics model for error correction through a corrected reference acceleration.
\newpage
\begin{align}
    \mathbf{\tau}_{\mathrm{PD}} &= \mathbf{K}_P(\mathbf{q}_\mathrm{d}-\mathbf{q}) + \mathbf{K}_D(\dot{\mathbf{q}}_\mathrm{d}-\dot{\mathbf{q}}) \label{eq:PD}\\
    \mathbf{\tau}_{\mathrm{PDG}} &= \mathbf{K}_P(\mathbf{q}_\mathrm{d}-\mathbf{q}) + \mathbf{K}_D(\dot{\mathbf{q}}_\mathrm{d}-\dot{\mathbf{q}}) + \mathbf{g}(\mathbf{q})\label{eq:PDG}\\
    \mathbf{\tau}_{\mathrm{PDFF}} &= \mathbf{K}_P(\mathbf{q}_\mathrm{d}-\mathbf{q}) + \mathbf{K}_D(\dot{\mathbf{q}}_\mathrm{d}-\dot{\mathbf{q}}) + \mathbf{f}_{\mathrm{dyn}}^{-1}(\mathbf{q}_{\mathrm{d}}, \dot{\mathbf{q}}_{\mathrm{d}}, \ddot{\mathbf{q}}_{\mathrm{d}}) \label{PDFF}\\ 
    \mathbf{\tau}_{\mathrm{ID}} &= \mathbf{f}_{\mathrm{dyn}}^{-1}(\mathbf{q}, \dot{\mathbf{q}}, \ddot{\mathbf{q}}_\mathrm{d} + \mathbf{K}_P(\mathbf{q}_\mathrm{d}-\mathbf{q}) + \mathbf{K}_D(\dot{\mathbf{q}}_\mathrm{d}-\dot{\mathbf{q}})) \label{eq:ID}
\end{align}

\vspace{5pt}
\section{Experiments}

As in our previous work, we use low restitution balls and cone-shaped unactuated hands, with all contacts modeled as spring-damper systems in MuJoCo~\cite{todorov2012mujoco}.
While this choice of a cone-shaped hand with $0.17$m diameter provides advantageous passive stability~\cite{ploeger2021high}, an actuated hand that fully opens at take-off can start moving sideways towards the following touch-down location earlier. This decreases the required vacant time $T_v$ and average number of balls in air per hand $W$.
In a simplified setup, we assume complete state control over two free-floating hands to verify the planned trajectories in task space independent of additional sources of errors like the inverse kinematics and tracking controllers.
In the setup depicted in figure~\ref{fig:title_pic}, the hands are attached to 4-DoF Barrett~WAM arms which are torque controlled at $500$Hz using the inverse dynamics controller~(\ref{eq:ID}) unless stated otherwise.
The robot bases are mounted jointly, shaping a torso with a shoulder width of $0.76$m. This configuration allows the elbows to drop down naturally and juggle in a human-typical pose, which is especially suitable for juggling since adaptations of the hand position in the horizontal catching plane only cause slight hand rotation.

While our method generalizes to fountain juggling, and we achieve stable juggling of six balls, the described unactuated hands are not suited for juggling even ball numbers requiring large carry distances $d_d$ to avoid ball collisions.
Therefore we restrict the presented experiments to cascade juggling patterns with odd numbers of balls.

In the following sections, we conduct several experiments to measure the performance of our approach and test the relevancy of individual components of planner and controller. Runs are started in running patterns for each experiment, with most balls already in the air.
Our primary metric to evaluate the stability of a juggling patter are the number of catches before the first ball drops, counting only balls that were thrown by the robot.
As we do not count balls that were initialized in air, this metric is comparable to human juggling, which typically starts with all balls held in hands.

\subsection{Verifying Theoretical Limits}
\begin{figure}
    \centering
    \vspace{5pt}
    \includegraphics[width=1\columnwidth, trim={0cm, 0cm , 0, 0}, clip]{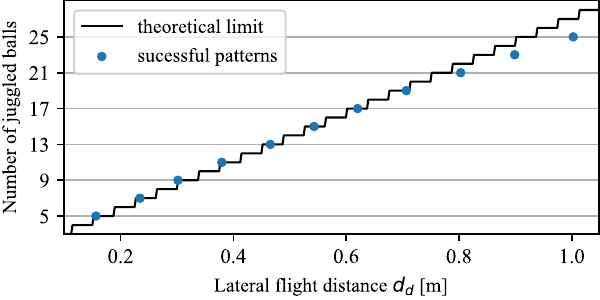}
    \caption{\textbf{Theoretical limit vs. achieved patterns:} The maximum number of balls is limited by the horizontal distance they travel. Stable juggling close to the upper bound was achieved in task space for more than $500$ catches.}
    \label{fig:theoretic_limit}
\end{figure}

The maximum amount of balls that can be juggled is limited by the distance between take-off and touch-down locations $d_f$ through the kinematic upper bound~(\ref{eq:limit}).
We validate this upper bound empirically in cascade patterns at dwell ratio $R=0.5$ using the simplified setup with floating hands.
As shown in figure~\ref{fig:theoretic_limit}, we achieve stable juggling for large amounts of balls up to 19 balls at the kinematic upper bound, defining stable as reaching $500$ catches. For $21$ and more balls at throw heights greater than $15$m, throwing precision decreases due to unmodeled contact dynamics, and extra lateral travel $d_f$ is required.
On the anthropomorphic two-arm setup, we maximally achieve $50.000$ catches with the maximum possible $17$ balls at ${d_f=0.62}$m, requiring balls to pass each other within a distance of $2.5$mm.

\subsection{Robustness to Disturbances}
\begin{figure}
    \centering
    \includegraphics[width=1\columnwidth, trim={0cm, 0cm , 0, 0}, clip]{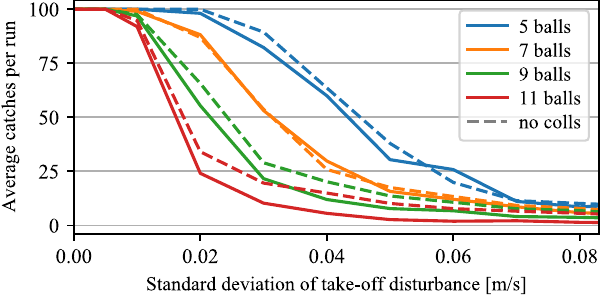}
    \caption{\textbf{Robustness to disturbances:} Ball trajectories in cascade patterns were disturbed by isotropic white noise on the take-off velocity. Catches counted until $100$ averaged over $50$ runs, with and without ball collisions.}
    \vspace{-10pt}
    \label{fig:disturbances}
\end{figure}

We disturbed each ball trajectory immediately after take-off by adding isotropic white noise to the velocity to evaluate the robustness of cascade patterns achieved with the two-arm setup.
For each data point in figure~\ref{fig:disturbances}, we counted the number of catches up to a maximum of 100 and averaged over $50$ trials.
As expected, higher numbers of balls are more susceptible to disturbances.
To identify what causes balls to drop, we reran the entire experiment with all inter ball collisions turned off---balls tunneling through each other---indicated by the dashed lines.
It is apparent that the importance of ball collisions increases with the number of balls, but their overall contribution to the degradation of robustness is only minor compared to the compounding turbulence caused by insufficiently dissipated kinetic energy.

\subsection{Necessity of Trajectory Constraints}
\begin{figure}
    \centering
    \vspace{5pt}
    \includegraphics[width=1\columnwidth, trim={0cm, 0cm , 0cm, 0}, clip]{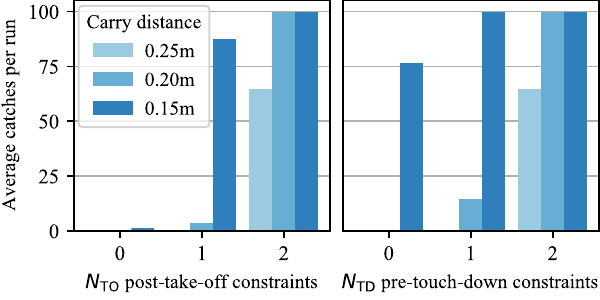}
    \caption{\textbf{Necessity of trajectory constraints:} Alternating the number of constrained support points $N_{\mathrm{TO}}$ and $N_{\mathrm{TO}}$ for three different carry distances $d_d$ in a five-ball cascade pattern shows the indispensability of the pre-touch-down and post-take-off constraints. The achieved catches were capped at $100$ and averaged over $50$ runs with take-off ball disturbance of $\sigma=0.01$m/s.}
    \label{fig:constraints}
\end{figure}

The additional constraints~(\ref{eq:pre_TD_vel}-\ref{eq:post_TO_kin}) are enforced on short intervals after take-off and before touch-down to prevent unplanned hand-ball contacts and ensure the successful execution of throws and catches. In the time-discrete case, these constraints are only implemented on the $N_{\mathrm{TO}}$ and $N_{\mathrm{TD}}$ support points within these intervals.
To evaluate if these additional constraints are necessary and on which interval they need to hold, we again count catches until a maximum of $100$, average over $50$ trials, and disturb ball velocities by the isotropic white noise $\sigma=0.01$m/s immediately after take-off.
We vary $N_{\mathrm{TO}}$ and $N_{\mathrm{TD}}$ between $0$~and~$2$ while keeping the other constant at $2$. Figure~\ref{fig:constraints} shows that juggling a five-ball cascade pattern is infeasible without these additional constraints and that they become more critical at more considerable carry distances.
Juggling with minimal carry distances, to avoid the additional constraints, would render the pattern susceptible to disturbances.

\begin{figure}
    \centering
    \includegraphics[width=1\columnwidth, trim={0cm, 0cm , 0, 0}, clip]{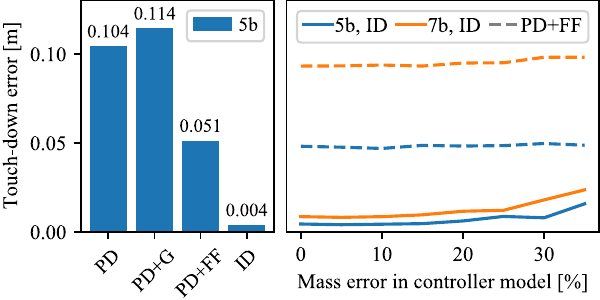}
    \vspace{-10pt}
    \caption{\textbf{Evaluating tracking performance:} Evaluating the average error of touch-down locations in undisturbed five-ball cascade patterns (Left) Comparison of different tracking controllers show the need for inverse dynamics control. (Right) While ID control tracks closer, PD+FF control is less susceptible to varying uncertainties in the dynamics model.}
    \vspace{-10pt}
    \label{fig:control}
\end{figure}

\subsection{Tracking Control}
We compared the tracking controllers~(\ref{eq:PD}-\ref{eq:ID}) by evaluating them on the throwing accuracy, averaging the distance between planned and achieved touch-down locations.
For the task of juggling five balls in an undisturbed cascade pattern, this average touch-down error is shown for each controller on the left side of figure~\ref{fig:control}.
We show that the simple PD and PD+G controllers are at a significant disadvantage by not taking the robot's inertia into account.
Juggling more than five balls with either of them was infeasible.
While adding a feed-forward torque, the PD+FF controller halves the touch-down error, the ID controller nearly eliminates it.
\newpage
Since the ground truth dynamics model of a physical system are never known, we tested the touch-down error of the PD+FF and the ID controller with respect to model errors, juggling five and seven balls.
We averaged the touch-down error of the first $20$ catches over $50$ different models with masses varied by standard deviations relative to the ground truth masses. The right side of figure~\ref{fig:control} shows that the overall average touch-down error of ID-controlled throws is substantially lower but more susceptible to model errors.

\section{Conclusions}

We presented a novel decomposition of the infinite-horizon toss juggling movement into a sequential short-horizon trajectory optimization problem.
Moreover, we identified the critical constraints necessary for dexterous toss juggling. These constraints generalize to other dynamic manipulation tasks with switching contacts.
To the best of our knowledge, this formulation is the first to demonstrate stable juggling of five and more objects with anthropomorphic manipulators in a physics based simulation, reaching the robot's kinematic upper bound of $17$ balls.

We plan to evaluate our approach on a real-world physical robot. In prior work on a physical platform, we reached stable juggling of two balls in one hand, equivalent to juggling four balls in a fountains pattern. Since we did not reach the robot's torque limits, a five-ball cascade pattern appears to be a realistic goal given the current hardware constraints.
Our current choice of end-effector design restricts possible lateral movements with an object and the possible types of objects due to the lack of active grasping. Therefore, it may also be worthwhile to further investigate different hand choices.
While in this work we focused on the stationary infinite-horizon control problem of running a juggling pattern, experiments on a physical system will additionally require a solution to the transient short-horizon control problem of starting a juggling pattern.

Our approach is limited by the assumptions in definition~\ref{def:toss_juggling}. 
The restriction to non-zero dwell times in a) separates toss juggling from the different skill of paddle juggling. However, the three other assumptions lead to exciting directions for future work.
Firstly, relaxing assumption b) and accounting for throws of different heights grants access to various juggling patterns mixing crossing and non-crossing throws.
Secondly, by relaxing assumption c) and adapting take-off and touch-down positions, online collision avoidance between ball trajectories can be taken into account in real-time.
Finally, breaking the constant juggling rhythm required in assumption d) and adapting the phase of the juggling pattern to one or more partners allows for interactive juggling---either by sharing a joint pattern with skilled partners or by having a novice partner add a ball to a running pattern by throwing it in, or removing a ball by snatching it out.

\section*{ACKNOWLEDGEMENTS}
We thank Joe Watson and the reviewers for their valuable feedback.
This work was supported by the Hessian research priority programme LOEWE within the project “WhiteBox”.

\addtolength{\textheight}{-5cm}   


\newpage
\bibliography{references}
\bibliographystyle{ieeeconf}

\end{document}